\documentclass[10pt,twocolumn,letterpaper]{article}

\usepackage{cvpr}
\usepackage{graphicx}
\usepackage{color}
\usepackage{times}
\usepackage{amsmath,mathtools}
\usepackage{amsthm}	
\usepackage{amssymb}
\usepackage{enumerate}
\usepackage{subfigure}
\usepackage{algorithm} 
\usepackage{algpseudocode}
\usepackage{booktabs}
\usepackage{tabularx}
\usepackage{array}
\usepackage{bm}
\usepackage{url}
\usepackage{bbm}
\usepackage{enumitem}
\usepackage{authblk}
\usepackage[pagebackref=true,breaklinks=true,letterpaper=true,colorlinks,bookmarks=false]{hyperref}

\usepackage{pifont}
\usepackage[symbol*]{footmisc}

\DefineFNsymbolsTM{otherfnsymbols}{%
  \textbullet \circ
  \textsection   \mathsection
  \textdagger    \dagger
  \textdaggerdbl \ddagger
  \textasteriskcentered *
  \textbardbl    \|%
  \textparagraph \mathparagraph
}%

\setfnsymbol{otherfnsymbols}
\newcommand{\cmark}{\ding{61}}%
\newcommand{\xmark}{*}%
\newcommand{\zmark}{\ding{67}}%

\cvprfinalcopy 

\def\cvprPaperID{****} 

\ifcvprfinal\fi

\newcommand{\bld}[1]{{\bf{#1}}}

\DeclareMathOperator{\atantwo}{atan2}

\newtheoremstyle{break}
  {\topsep}{\topsep}%
  {\itshape}{}%
  {\bfseries}{}%
  {\newline}{}%

\theoremstyle{break}

\newcommand{\insertimageC}[5]{ 
\begin{figure}[#5]
\centering
\includegraphics[width=#1\linewidth, clip=true]{figures/#2}
\caption{#3}
\label{#4}
\vspace{-2mm}
\end{figure}
}

\newcommand{\insertimageStar}[5]{ 
\begin{figure*}[#5]
\centering
\includegraphics[width=#1\linewidth, clip=true]{figures/#2}
\caption{#3}
\label{#4}
\vspace{-2mm}
\end{figure*}
}

\algnewcommand\algorithmicinput{\textbf{Input:}}
\algnewcommand\INPUT{\item[\algorithmicinput]}
\algnewcommand\algorithmicoutput{\textbf{Output:}}
\algnewcommand\OUTPUT{\item[\algorithmicinput]}

\newcommand{\comment}[1]{}

\newcommand{\suchthat}{\;\ifnum\currentgrouptype=16 \middle\fi|\;}

\begin{document}

\title{PPFNet: Global Context Aware Local Features for Robust 3D Point Matching}

\author[\cmark,\xmark,\zmark]{Haowen Deng}
\author[\cmark,\xmark]{Tolga Birdal}
\author[\cmark,\xmark]{Slobodan Ilic}
\affil[\cmark]{ Technische Universitat M\"{u}nchen, Germany,\quad \xmark~Siemens AG, M\"{u}nchen, Germany}
\affil[\zmark]{ National University of Defense Technology, China\\

{\normalsize \tt{haowen.deng@tum.de\,}},  {\normalsize \tt{\,tolga.birdal@tum.de\,}}, {\normalsize \tt{\,slobodan.ilic@siemens.com}}}

\maketitle

\begin{abstract}
We present PPFNet - Point Pair Feature NETwork for deeply learning a globally informed 3D local feature descriptor to find correspondences in unorganized point clouds. PPFNet learns local descriptors on pure geometry and is highly aware of the global context, an important cue in deep learning. Our 3D representation is computed as a collection of point-pair-features combined with the points and normals within a local vicinity. Our permutation invariant network design is inspired by PointNet and sets PPFNet to be ordering-free. As opposed to voxelization, our method is able to consume raw point clouds to exploit the full sparsity. PPFNet uses a novel \textit{N-tuple} loss and architecture injecting the global information naturally into the local descriptor. It shows that context awareness also boosts the local feature representation. Qualitative and quantitative evaluations of our network suggest increased recall, improved robustness and invariance as well as a vital step in the 3D descriptor extraction performance.
\end{abstract}

\section{Introduction}
Local description of 3D geometry plays a key role in 3D vision as it precedes fundamental tasks such as correspondence estimation, matching, registration, object detection or shape retrieval. Such wide application makes the local features amenable for use in robotics~\cite{choi2012voting}, navigation (SLAM)~\cite{salas2013slam++} and scene reconstruction for creation of VR contents and digitalization. Developing such a general-purpose tool motivated scholars to hand-craft their 3D feature descriptors/signatures for decades~\cite{salti2014shot,Rusu08IAS,rusu2009fast}. Unfortunately, we now notice that this quest has not been very fruitful in generating the desired repeatable and discriminative local descriptors for 3D point cloud data, especially when the input is partial or noisy~\cite{guo2014performance,kiforenko2017performance}. 

The recent trends carefully follow the paradigm shift to deep neural networks, but the latest works either base the representation on a hand-crafted input encoding~\cite{Khoury_2017_ICCV} or try to naively extend the networks from 2D domain to 3D~\cite{zeng20163dmatch}. Both approaches are sub-optimal, as they do not address an end-to-end learning of the raw data, point sets.

\insertimageC{1}{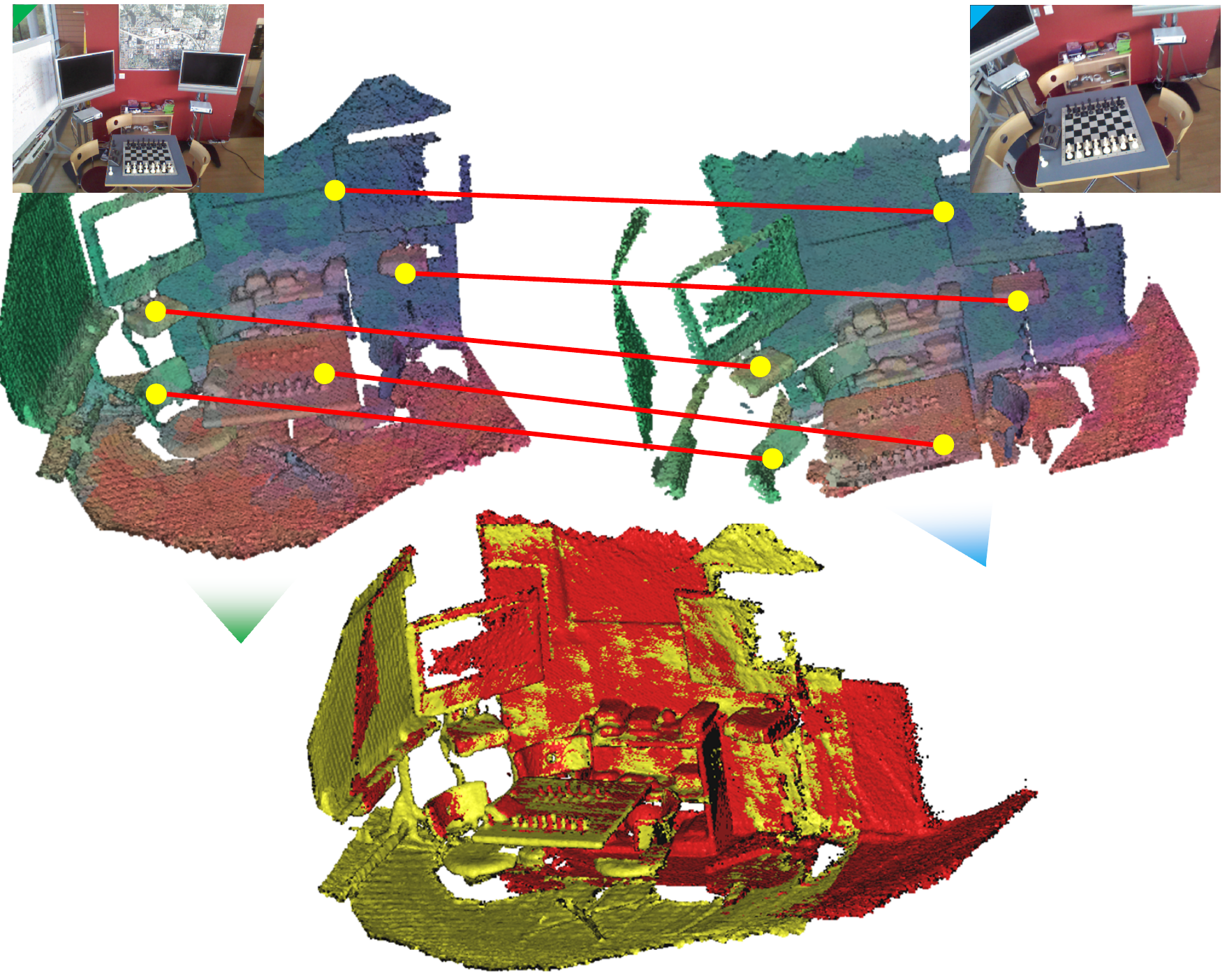}{PPFNet generates repeatable, discriminative descriptors and can discover the correspondences simultaneously given a pair of fragments.  Point sets are colored by a low dimensional embedding of the local feature for visualization. 3D data and the illustrative image are taken from 7-scenes dataset~\cite{shotton2013scene}.}{fig:teaser}{t!}

In this paper, we present \textit{PPFNet} network for deep learning of fast and discriminative 3D local patch descriptor with increased tolerance to rotations. To satisfy its desirable properties, we first represent the local geometry with an augmented set of simple geometric relationships: points, normals and point pair features (PPF)~\cite{rusu2009fast,drost2010model}. We then design a novel loss function, which we term as \textit{N-tuple loss}, to simultaneously embed multiple matching and non-matching pairs into a Euclidean domain. Our loss resembles the contrastive loss~\cite{hadsell2006dimensionality}, but instead of pairs, we consider an N-combination of the input points within two scene fragments to boost the separability. Thanks to this many-to-many loss function, we are able to inject the global context into the learning, i.e. PPFNet is aware of the other local features when establishing correspondence for a single one. Also because of such parallel processing, PPFNet is very fast in inference. Finally, we combine all these contributions in a new pipeline, which trains our network from correspondences in 3D fragment pairs. PPFNet extends PointNet~\cite{qi2016pointnet} and thereby is natural for point clouds and neutral to permutations. Fig.~\ref{fig:teaser} visualizes our features and illustrates the robust matching we can perform. 

Our extensive evaluations show that PPFNet achieves the state of the art performance in accuracy, speed, robustness to point density and tolerance to changes in 3D pose.

\insertimageStar{1}{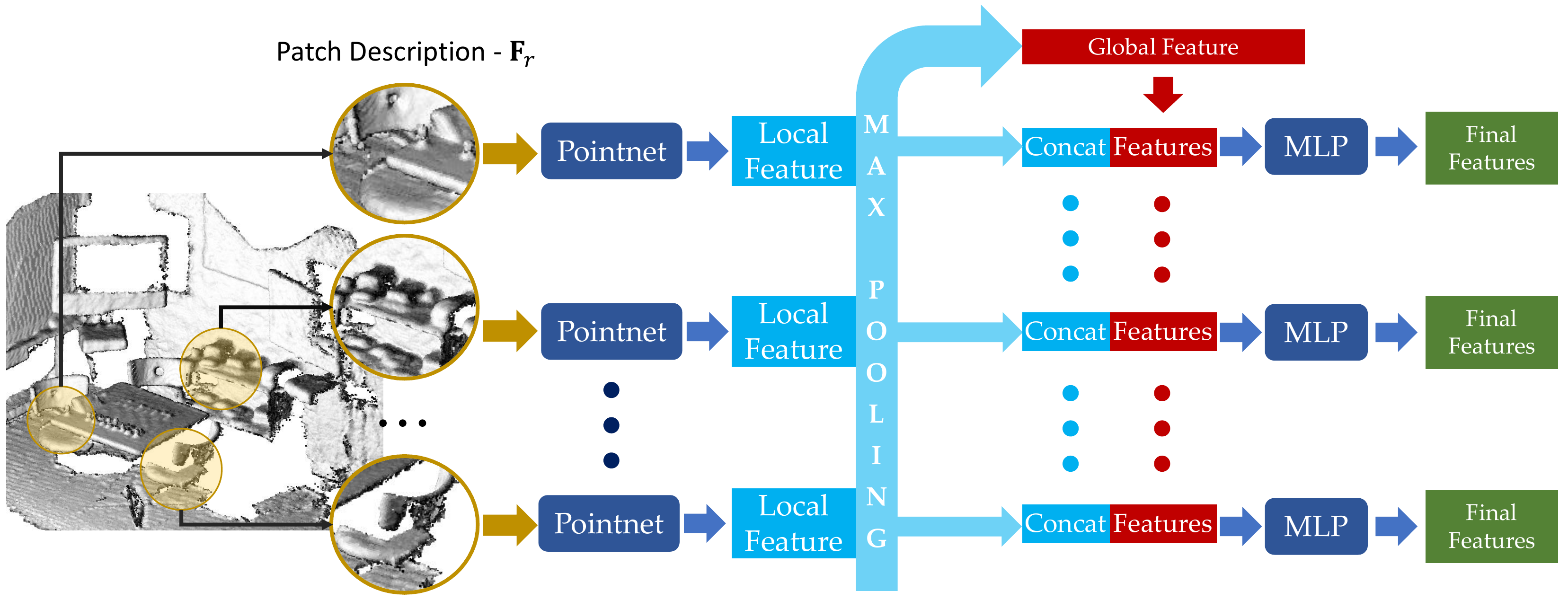}{PPFNET, our inference network, consists of multiple PointNets, each responsible for a local patch. To capture the global context across all local patches, we use a max-pooling aggregation and fusing the output back into the local description. This way we are able to produce stronger and more discriminative local representations.}{fig:ppfnet}{t!}
\section{Related Work}
\label{sec:related}
\paragraph{Hand-crafted 3D Feature Descriptors}
Similar to its 2D counterpart, extracting meaningful and robust local descriptors from 3D data kept the 3D computer vision researchers busy for a long period of time~\cite{salti2014shot,rusu2009fast,tombari2010unique,johnson1999using,guo2013rops}. Unfortunately, contrary to 2D, the repeatability and distinctiveness of 3D features were found to be way below expectations~\cite{guo2016comprehensive,guo2014performance,kiforenko2017performance}. Many of those approaches try to discover a local reference frame (LRF), which is by its simplest definition, non-unique~\cite{guo2013rotational}. This shifts the attention to LRF-free methods such as the rotation invariant point pair features (PPF) to be used as basis for creating powerful descriptors like PPFH~\cite{Rusu08IAS} and FPFH ~\cite{rusu2009fast}. PPFs are also made semi-global to perform reasonably well under difficult scenarios, such as clutter and occlusions ~\cite{drost2010model,birdal3dv2015,hinterstoisser2016going,birdal2017cad}. Thus, they have been applied in many problems to estimate 3D poses or retrieve and recognize objects~\cite{rusu2009fast,wahl2003surflet}. Thanks to the simplicity and invariance properties of PPFs, along with the raw points and normals, we use it in PPFNet to describe the local geometry and learn a strong local descriptor.

\paragraph{Learned 3D Feature Descriptors}
With the advent of deep learning, several problems like 3D retrieval, recognition, segmentation and descriptor learning have been addressed using 3D data~\cite{wu20153d,socher2012convolutional}. A majority of them operate on depth images~\cite{kehl2016deep,wohlhart2015learning}, while just a handful of works target point clouds directly. As we address the problem of 3D descriptor learning on point sets, we opt to review the data representations regardless the application and then consider learning local descriptors for point cloud matching.

There are many ways to represent sparse unstructured 3D data. Early works exploited the apparent dense voxels ~\cite{maturana2015voxnet,qi2016volumetric}, usually in form of a binary-occupancy grid. This idea is quickly extended to more informative encoding such as TDF~\cite{song2016deep}, TSDF~\cite{zeng20163dmatch} (Truncated Signed Distance Field), multi-label occupancy~\cite{wu20153d} and other different ones~\cite{huang2016point,yarotsky2017geometric}. Since mainly used in the context of 3D retrieval, entire 3D objects were represented with small voxel grids $30^3$ limited by the maximal size of 3D convolutions kernels~\cite{maturana2015voxnet}. These representations have also been used for describing the local neighbours in the context of 3D descriptor learning. One such contemporary work is 3DMatch ~\cite{zeng20163dmatch}. It is based on a robust volumetric TSDF encoding with a contrastive loss to learn correspondences. Albeit straightforward, 3DMatch ignores the raw nature of the input: sparsity and unstructured-ness. It uses dense local grids and 3D CNNs to learn the descriptor and thus can fall short in training/testing performance and recall. 

A parallel track of works follow a view based scheme~\cite{Elbaz_2017_CVPR,huangKCCKY17}, where the sub-spaces of 3D information in form of projections or depth map are learned with well studied 2D networks. Promising potential, these methods do not cover for sparse point sets. Another spectrum of research exploits graph networks~\cite{kipf2016semi,defferrard2016convolutional} also to represent point sets~\cite{Qi_2017_ICCV}. This new direction prospers in other domains but graph neural networks are not really suited to point clouds, as they require edges, not naturally arising from point sets. Khoury et. al.~\cite{Khoury_2017_ICCV} try to overcome the local representation problem by a hand-crafted approach and use a deep-network only for a dimensionality reduction. Their algorithm also computes the non-unique LRF and taking such a path deviates from the efforts of end-to-end learning.

The low hanging vital step is taken by PointNet~\cite{qi2016pointnet}, a network designed for raw 3D point input. PointNet demonstrated that neural networks can be designed in a permutation invariant manner to learn segmentation, classification or keypoint extraction. It is then extended to PointNet++ to better handle the variations in point density~\cite{qi2017pointnet++}. However, this work in its original form, cannot tackle the problem of local geometry description as we successfully do.

As we will describe, PPFNet, trained on PPFs, points and normals of local patches, boosts the notion of global context in the semi-local features of PointNet by optimizing combinatorial matching loss between multitudes of patches, resulting in powerful features, outperforming the prior art.

\section{Background}
\paragraph{Motivation}
Before explaining our local descriptor, let us map out our setting.
Consider two 3D point sets $\bld{X} \in \mathbb{R}^{n \times 3}$ and $\bld{Y} \in \mathbb{R}^{n \times 3}$. Let $\bld{x}_i$ and $\bld{y}_i$ denote the coordinates of the $\text{i}^{\text{th}}$ points in the two sets, respectively. Momentarily, we assume that for each $\bld{x}_i$, there exists a corresponding $\bld{y}_i$, a bijective map. Then following \cite{li20073d} and assuming rigidity, the two sets are related by a correspondence and pose (motion), represented by a permutation matrix $\bld{P}\in \mathbb{P}^n$ and a rigid transformation $\bld{T}=\{\bld{R} \in SO(3), \bld{t}\in \mathbb{R}^3\}$, respectively. Then the $L_2$ error of point set registration reads:
\begin{equation}
d(\bld{X}, \bld{Y} | \bld{R}, \bld{t}, \bld{P}) = \frac{1}{n} \sum\limits_{i=1}^n \| \bld{x}_i - \bld{R}\bld{y}_{i(\bld{P})} - \bld{t} \|^2
\end{equation}
where $\bld{x}_i$ and $\bld{y}_{i(\bld{P})}$ are matched under $\bld{P}$.
We assume $\bld{X}$ and $\bld{Y}$ are of equal cardinality ($|\bld{X}|=|\bld{Y}|=n$). 
In form of homogenized matrices, the following is equivalent:
\begin{equation}
d(\bld{X}, \bld{Y} | \bld{T}, \bld{P}) = \frac{1}{n} \| \bld{X} - \bld{P} \bld{Y} \bld{T}^T \|^2
\end{equation}
Two sets are ideally matching if $d(\bld{X}, \bld{Y} | \bld{T}, \bld{P})\approx 0$. This view suggests that, to learn an effective representation, one should preserve a similar distance in the embedded space:
\begin{equation}
d_f(\bld{X}, \bld{Y} | \bld{T}, \bld{P}) = \frac{1}{n} \| f(\bld{X}) - f(\bld{P} \bld{Y} \bld{T}^T) \|^2
\end{equation}
and $d_f(\bld{X}, \bld{Y} | \bld{T}, \bld{P})\approx 0 $ also holds for matching points sets under any action of $(\bld{T}, \bld{P})$. Thus, for invariance, it is desirable to have: $f(\bld{Y}) \approx f(\bld{P} \bld{Y} \bld{T}^T)$. Ideally we would like to learn $f$ being invariant to permutations $\bld{P}$ and as intolerant as possible to rigid transformations $\bld{T}$. Therefore, in this paper we choose to use a minimally handcrafted point set to deeply learn the representation. This motivates us to exploit PointNet architecture~\cite{qi2016pointnet} which intrinsically accounts for unordered sets and consumes sparse input. To boost the tolerance to transformations, we will benefit from point-pair-features, the true invariants under Euclidean isometry.
\vspace{-3mm}
\paragraph{Point Pair Features (PPF)}
Point pair features are antisymmetric 4D descriptors, describing the surface of a pair of oriented 3D points $\bld{x}_1$ and $\bld{x}_2$, constructed as:
\begin{equation}
\bm{\psi}_{12} = (\Vert \bld{d}\rVert_2, \angle(\bld{n}_1,\bld{d}), \angle(\bld{n}_2,\bld{d}), \angle(\bld{n}_1,\bld{n}_2))
\end{equation}
where $\bld{d}$ denotes the difference vector between points, $\bld{n}_1$ and $\bld{n}_2$ are the surface normals at $\bld{x}_1$ and $\bld{x}_2$. $\lVert \cdot \rVert$ is the Euclidean distance and $\angle$ is the angle operator computed in a numerically robust manner as in \cite{birdal3dv2015}:
\begin{equation}
\label{eq:angle}
\angle(\bld{v}_1,\bld{v}_2) = \atantwo \big( { \Vert \bld{v}_1 \times \bld{v}_2 \rVert} \, , \, {\bld{v}_1 \cdot \bld{v}_2}\big)
\end{equation}
$\angle(\bld{v}_1,\bld{v}_2)$ is guaranteed to lie in the range $[0,\pi)$. By construction, this feature is invariant under Euclidean transformations and reflections as the distances and angles are preserved between every pair of points. 
\vspace{-3mm}
\paragraph{PointNet}
PointNet~\cite{qi2016pointnet} is an inspiring pioneer addressing the issue of consuming point clouds within a network architecture. It is composed of stacking independent MLPs anchored on points up until the last layers where a high dimensional descriptor is synthesized. This descriptor is weak and used in max-pooling in order to aggregate to a global information, which is then fed into task specific losses. Use of the max-pooling function makes the network inconsiderate of the input ordering and that way extends notions of deep learning to point sets. It showed potential on tasks like 3D model classification and segmentation. Yet, local features of PointNet are only suitable for the tasks it targets and are not generic. Moreover, the spatial transformer layer employed can bring only marginal improvement over the basic architectures. It is one aspect of PPFNet to successfully cure these drawbacks for the task of 3D matching. Note, in our work, we use the vanilla version of PointNet.

\section{PPFNet}
\paragraph{Overview} We will begin by explaining our \textit{input preparation} to compute a set describing the local geometry of a 3D point cloud. We then elaborate on \textit{PPFNet architecture}, which is designed to process such data with merit. Finally, we explain our training method with a new loss function to solve the combinatorial correspondence problem in a global manner. In the end, the output of our network is a local descriptor per each sample point as shown in Fig.~\ref{fig:ppfnet}.

\insertimageC{1}{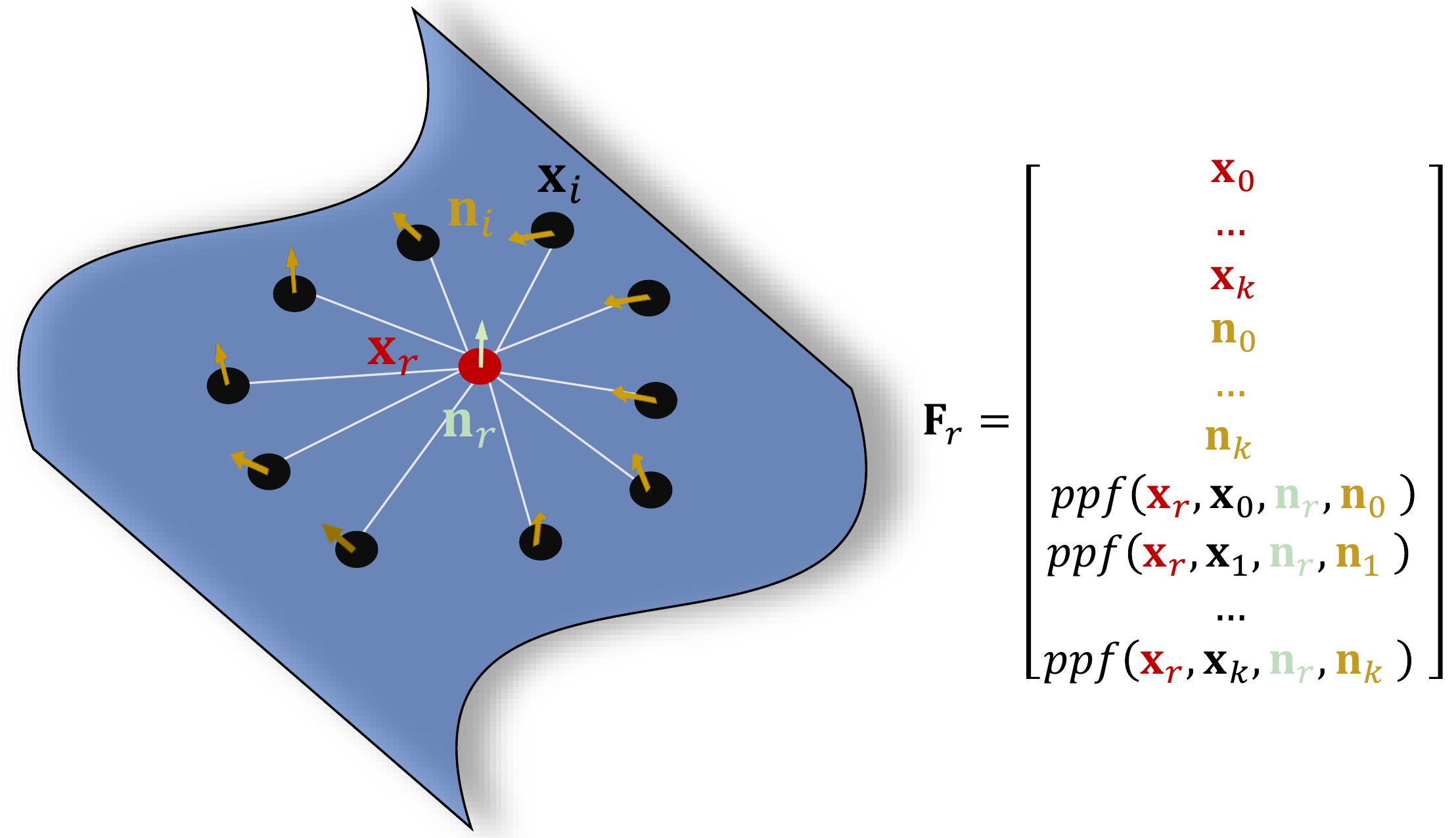}{Simplistic encoding of a local patch.}{fig:ppf}{t!}
\vspace{-3mm}
\paragraph{Encoding of Local Geometry}
Given a reference point lying on a point cloud $\bld{x}_r \in \textbf{X}$, we define a local region $\bld{\Omega} \subset \textbf{X}$ and collect a set of points $\{\bm{m}_i\} \in \bld{\Omega}$ in this local vicinity. We also compute the normals of the point set~\cite{Hoppe1992}. The associated local reference frame~\cite{tombari2010unique} then aligns the patches with canonical axes. Altogether, the oriented $\{\bm{x}_r \cup \{\bm{x}_i\}\}$ represent a local geometry, which we term \textit{a local patch}. We then pair each neighboring point $i$ with the reference $r$ and compute the PPFs. Note that complexity-wise, this is indifferent than using the points themselves, as we omit the quadratic pairing thanks to fixation of central reference point $\bld{x}_r$. As shown in Fig.~\ref{fig:ppf}, our final local geometry description and input to PPFNet is a combined set of points normals and PPFs:
\begin{equation}
\mathbf{F}_r = \{\mathbf{x}_r, \mathbf{n}_r, \mathbf{x}_i, \cdots, \mathbf{n}_i, \cdots, \bm{\psi}_{ri}, \cdots\}    
\end{equation}
\vspace{-10mm}
\paragraph{Network architecture}
The overall architecture of PPFNet is shown in Fig. \ref{fig:ppfnet}.  Our input consists of $N$ local patches uniformly sampled from a fragment. Due to sparsity of point-style data representation and efficient GPU utilization of PointNet, PPFNet can absorb those $N$ patches concurrently. The first module of PPFNet is a group of mini-PointNets, extracting features from local patches. Weights and gradients are shared across all PointNets during training. A max pooling layer then aggregates all the local features into a global one, summarizing the distinct local information to the global context of the whole fragment. This global feature is then concatenated to every local feature. A group of MLPs are used to further fuse the global and local features into the final global-context aware local descriptor.
\vspace{-3mm}
\paragraph{N-tuple loss}
Our goal is to use PPFNet to extract features for local patches, a process of mapping from a high dimensional non-linear data space into a low dimensional linear feature space. Distinctiveness of the resulting features are closely related to the separability in the embedded space. Ideally, the proximity of neighboring patches in the data space should be preserved in the feature space.

To this end, the state of the art seems to adopt two loss functions: contrastive~\cite{zeng20163dmatch} and triplet~\cite{Khoury_2017_ICCV}, which try to consider pairs and triplets respectively. Yet, a fragment consists of more than 3 patches and in that case the widely followed practice trains networks by randomly retrieving 2/3-tuples of patches from the dataset. However, networks trained in such manner only learn to differentiate maximum 3 patches, preventing them from uncovering the true matching, which is combinatorial in the patch count. 

Generalizing these losses to N-patches, we propose N-tuple loss, an $N$-to-$N$ contrastive loss, to correctly learn to solve this combinatorial problem by catering for the many-to-many relations as depicted in Fig.~\ref{fig:ntuples}. Given the ground truth transformation $\bld{T}$, 
N-tuple loss operates by constructing a correspondence matrix $\bld{M} \in \mathbb{R}^{N\times N}$ on the points of the aligned fragments. $\bld{M} = (m_{ij})$ where:
\begin{equation}
 m_{ij} = \mathbbm{1}(\lVert\bld{x}_i - \bld{T}\bld{y}_j \rVert_2 < \tau)
\end{equation}
$\mathbbm{1}$ is an indicator function. Likewise, we compute a feature-space distance matrix $\bld{D} \in \mathbb{R}^{N\times N}$ and $\bld{D} = (d_{ij})$ where
\begin{align}
d_{ij} = \lVert f(\bld{x}_i) - f(\bld{y}_j) \rVert_2
\end{align}
\insertimageC{1}{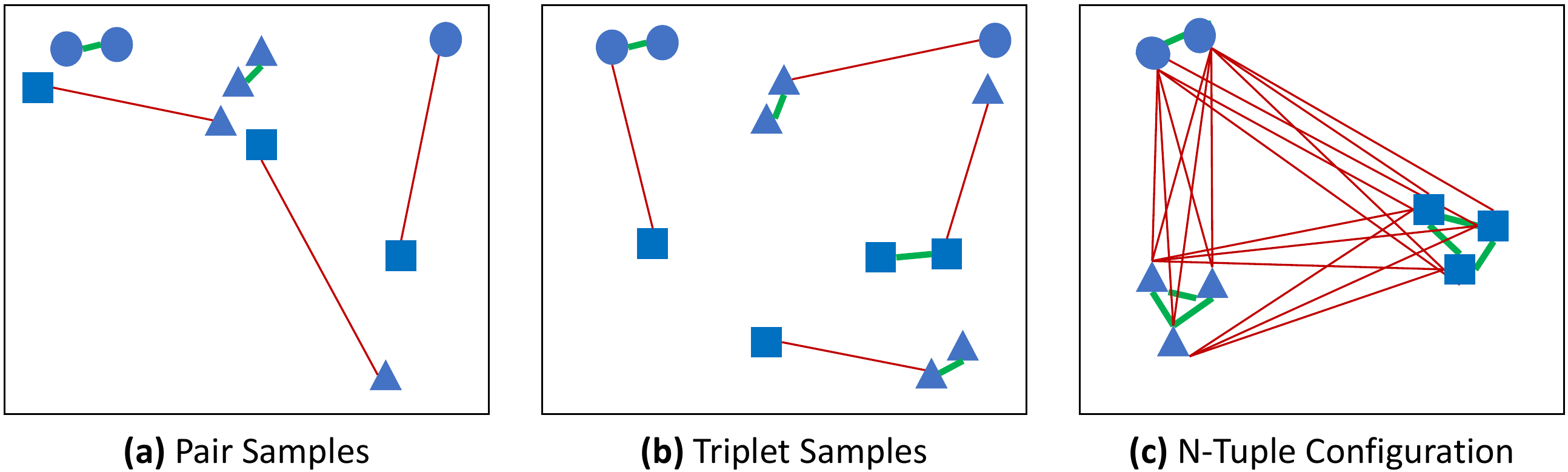}{Illustration of N-tuple sampling in feature space. Green lines link similar pairs, which are coerced to keep close. Red lines connect non-similar pairs, pushed further apart. Without N-tuple loss, there remains to be some non-similar patches that are close in the feature space and some distant similar patches. Our novel N-tuple method pairs each patch with all the others guaranteeing that all the similar patches remain close and non-similar ones, distant.}{fig:ntuples}{t!}
\insertimageStar{1}{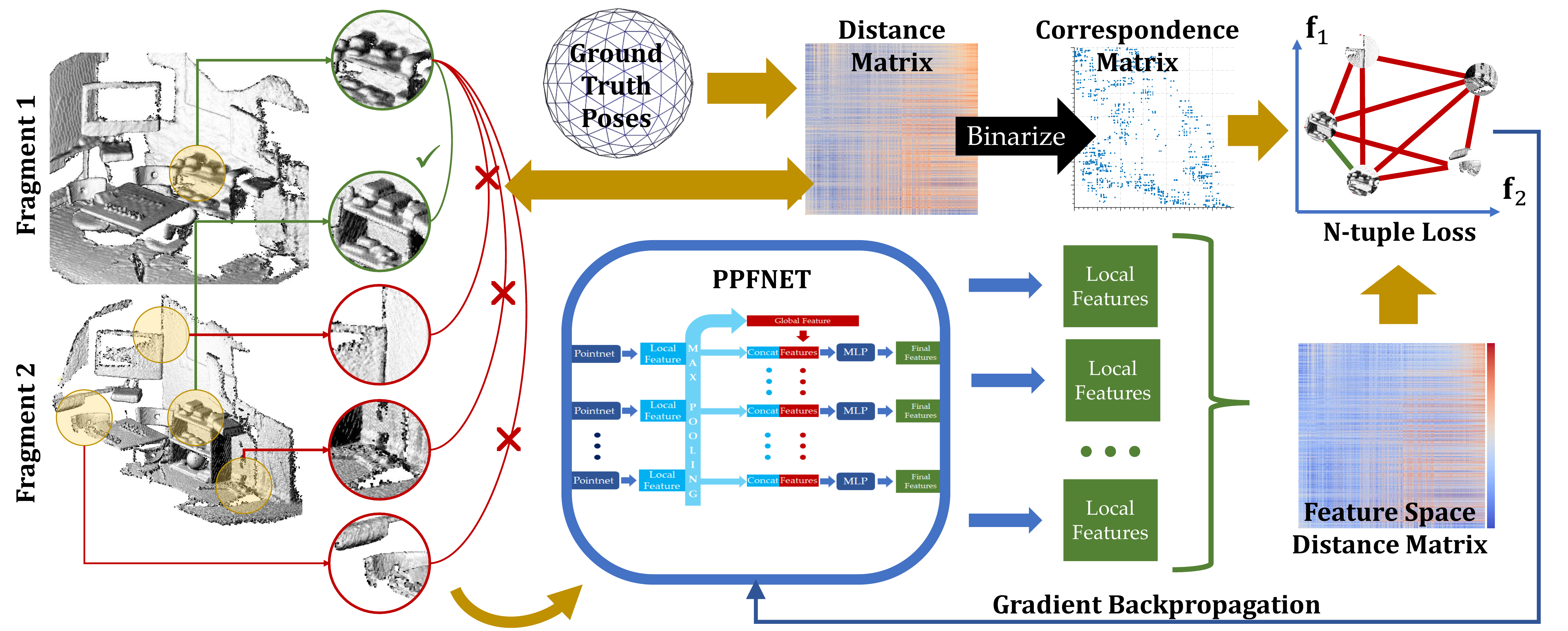}{Overall training pipeline of PPFNet. {Local patches are sampled from a pair of fragments respectively, and feed into PPFNet to get local features. Based on these features a feature distance matrix is computed for all the patch pairs. Meanwhile, a distance matrix of local patches is formed based on the ground-truth rigid pose between the fragments. By binarizing the distance matrix, we get a correspondence matrix to indicate all the matching and non-matching relationships between patches. N-tuple loss is then calculated by coupling the feature distance matrix and correspondence matrix to guide the PPFNet to find an optimal feature space.}}{fig:ppfnet_train}{t!}
\begin{table*}[t!]
\vspace{2mm}
  \centering
  \caption{Our evaluations on the 3DMatch benchmark before RANSAC. \textit{Kitchen} is from 7-scenes~\cite{shotton2013scene} and the rest from SUN3D~\cite{xiao2013sun3d}.}
  \resizebox{\textwidth}{!}{
    \begin{tabular}{lccccccccc}
    \toprule
          & Spin Images~\cite{johnson1999using} & SHOT~\cite{salti2014shot}  & FPFH~\cite{rusu2009fast}  & USC~\cite{tombari2010unique}   &
          PointNet~\cite{qi2016pointnet} &
          CGF~\cite{Khoury_2017_ICCV} &
          3DMatch~\cite{zeng20163dmatch} & 3DMatch-2K~\cite{zeng20163dmatch} &
          PPFNet (ours) \\
    \midrule
    Kitchen & 0.1937 & 0.1779 & 0.3063 & 0.5573 & 0.7115 & 0.4605 & 0.5751 & 0.5296 & \textbf{0.8972} \\
    Home 1  & 0.3974 & 0.3718 & 0.5833 & 0.3205 & 0.5513 & 0.6154 & \textbf{0.7372} & 0.6923 & 0.5577 \\
    Home 2  & 0.3654 & 0.3365 & 0.4663 & 0.3077 & 0.5385 & 0.5625 & \textbf{0.7067} & 0.6202 & 0.5913 \\
    Hotel 1 & 0.1814 & 0.2080 & 0.2611 & 0.5354 & 0.4071 & 0.4469 & 0.5708 & 0.4779 & \textbf{0.5796} \\
    Hotel 2 & 0.2019 & 0.2212 & 0.3269 & 0.1923 & 0.2885 & 0.3846 & 0.4423 & 0.4231 & \textbf{0.5769} \\
    Hotel 3 & 0.3148 & 0.3889 & 0.5000 & 0.3148 & 0.3333 & 0.5926 & \textbf{0.6296} & 0.5185 & 0.6111 \\
    Study   & 0.0548 & 0.0719 & 0.1541 & 0.5068 & 0.4315 & 0.4075 & \textbf{0.5616} & 0.3904 & 0.5342 \\
    MIT Lab & 0.1039 & 0.1299 & 0.2727 & 0.4675 & 0.5065 & 0.3506 & 0.5455 & 0.4156 & \textbf{0.6364} \\
    \midrule
    Average & 0.2267 & 0.2382 & 0.3589 & 0.4003 & 0.4710 & 0.4776 & 0.5961 & 0.5085 & \textbf{0.6231} \\
    \bottomrule
    \end{tabular}%
  \label{tab:3dmatchbenchmark}%
  }
\end{table*}%
The N-tuple loss then functions on the two distance matrices solving the correspondence problem. For simplisity of expression, we define an operation $\sum^{*}(\cdot)$ to sum up all the elements in a matrix. N-tuple loss can be written as:
\begin{equation}
\label{eq:n_tuple_loss}
\begin{split}
L = \sum\nolimits_{}^{*}\left ( \frac{\bld{M} \circ \bld{D}}{\lVert \bld{M} \rVert_2^2} + \alpha \frac{max(\theta - (1- \bld{M})\circ \bld{D}, 0)}{N^2 - \lVert \bld{M} \rVert_2^2} \right)
\end{split}
\end{equation}
Here $\circ$ stands for Hadamard Product - element-wise multiplication.  $\alpha$ is a hyper-parameter balancing the weight between matching and non-matching pairs and $\theta$ is the lower-bound on the expected distance between non-correspondent pairs. 
We train PPFNet via N-tuple loss, as shown in Fig. \ref{fig:ppfnet_train}, by drawing random pairs of fragments instead of patches. This also eases the preparation of training data. 
\section{Results}
\paragraph{Setup}
Our input encoding uses a 17-point neighborhood to compute the normals for the entire scene, using the well accepted plane fitting \cite{Hoppe1992}. For each fragment, we anchor 2048 sample points distributed with spatial uniformity. These sample points act as keypoints and within their $30cm$ vicinity, they form the patch, from which we compute the local PPF encoding. Similarly, we down-sample the points within each patch to 1024 to facilitate the training as well as to increase the robustness of features to various point density and missing part. For occasional patches with insufficient points in the defined neighborhood, we randomly repeat points to ensure identical patch size. PPFNet extracts compact descriptors of dimension 64.

PPFNet is implemented in the popular Tensorflow \cite{abadi2016tensorflow}. The initialization uses random weights and ADAM \cite{kingma2014adam} optimizer minimizes the loss. Our network operates simultaneously on all 2048 patches. Learning rate is set at 0.001 and exponentially decayed after every 10 epochs until 0.00001. Due to the hardware constraints, we use a batch size of 2 fragment pairs per iteration, containing 8192 local patches from 4 fragments already. This generates $2\times 2048^2$ combinations for the network per batch.
\begin{figure*}[t!]
\centering
\subfigure[Recall on 3DMatch benchmark]{
\includegraphics[width=0.2412\linewidth, clip=true]{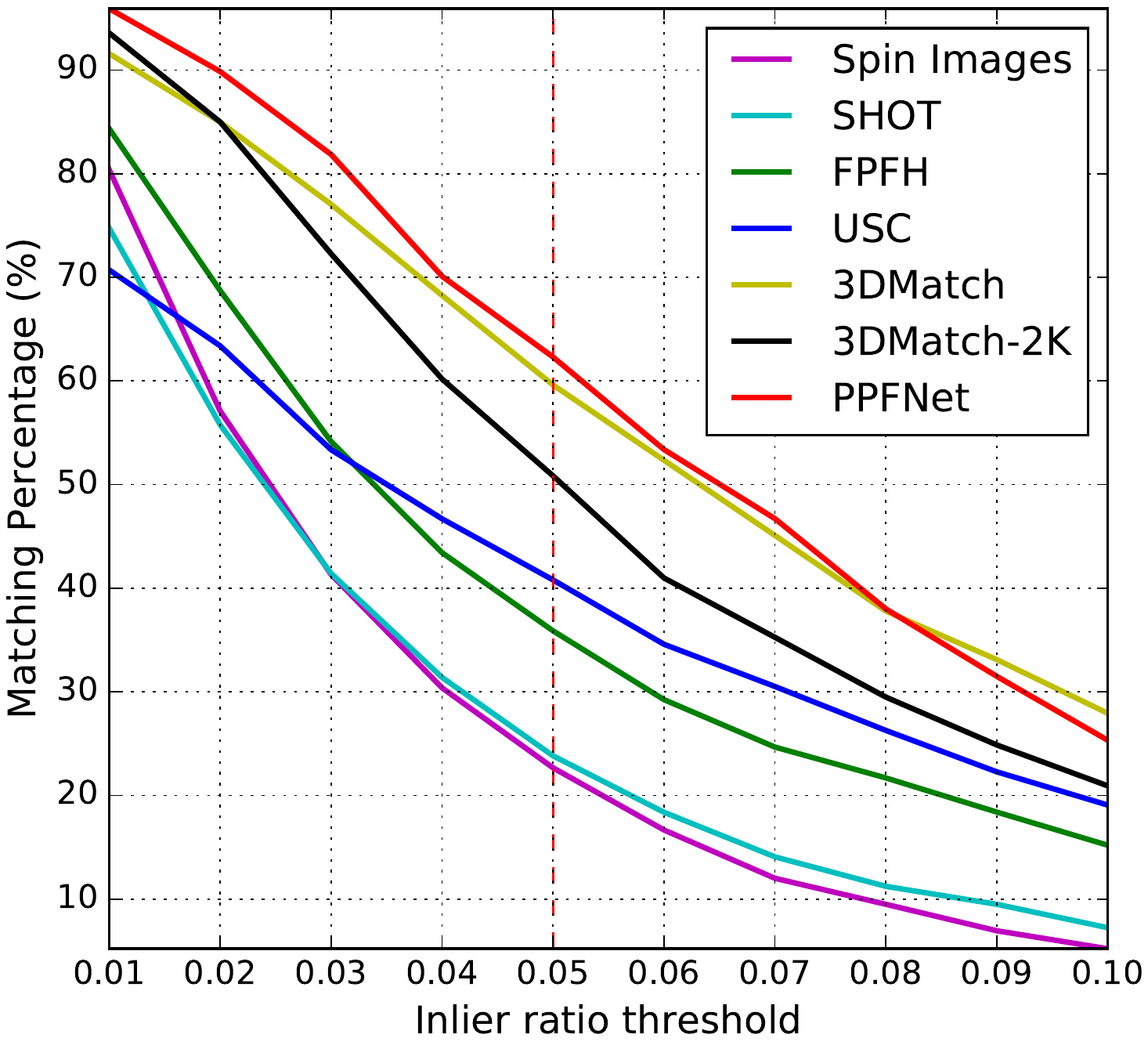}
\label{subfig:accuracy}}
\subfigure[Robustness to sparsity]{
\includegraphics[width=0.243\linewidth, clip=true]{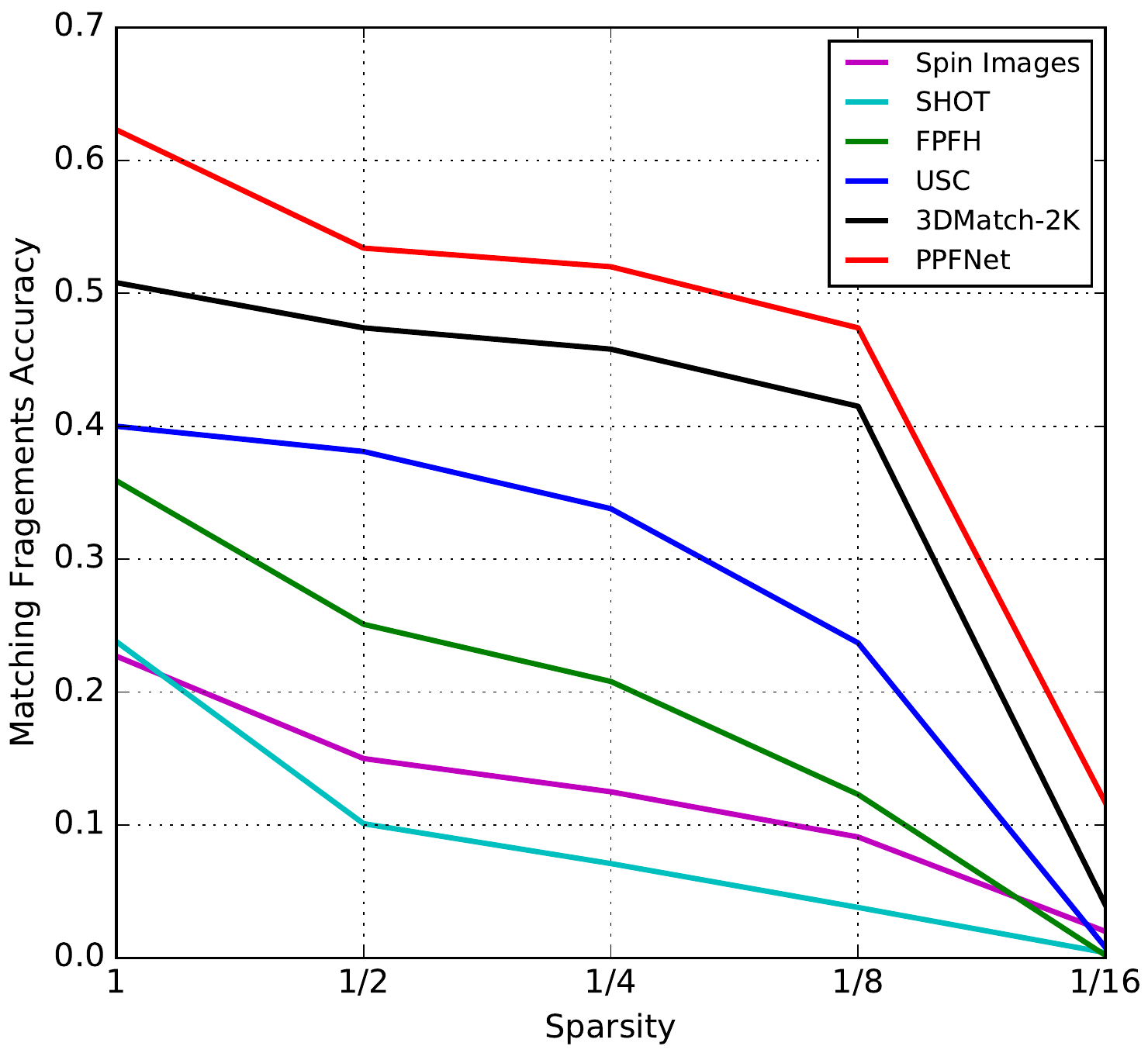}
\label{subfig:sparsity}}
\subfigure[Training with different inputs]{
\includegraphics[width=0.2305\linewidth, clip=true]{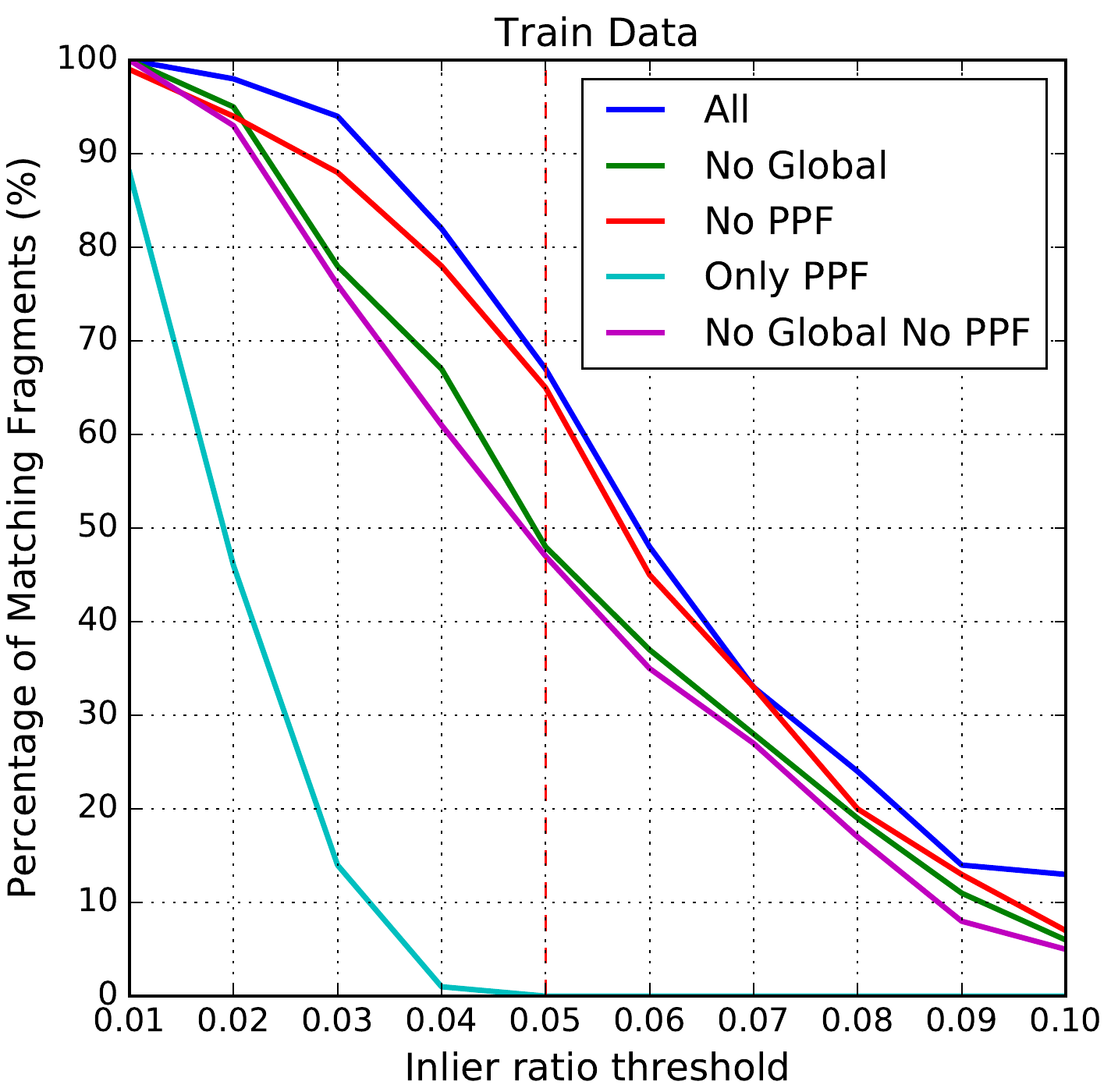}
\label{subfig:trainarch}}
\subfigure[Testing with different inputs]{
\includegraphics[width=0.2305\linewidth, clip=true]{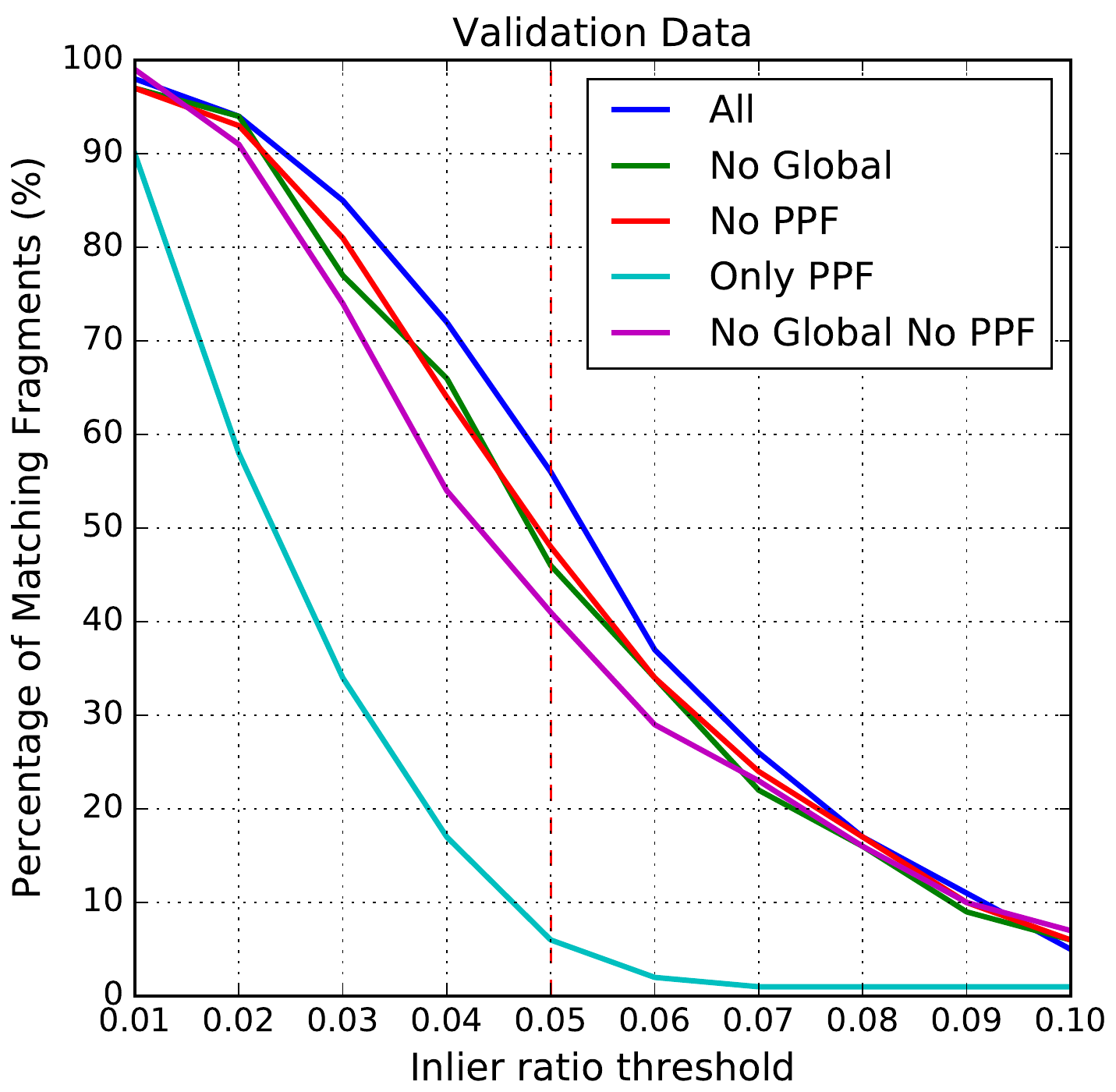}
\label{subfig:testarch}}
\caption{Evaluating PPFNet on real datasets: \textbf{(a)} Our method consistently outperforms the state-of-the-art on matching task (no RANSAC is used) in terms of recall. \textbf{(b)} Thanks to its careful design, PPFNet clearly yields the highest robustness to change in the sparsity of the input, even when only $6.25\%$ of the input data is used. \textbf{(c, d)} Assessing different elements of the input on training and validation sets, respectively. Note that combining cues of global information and point pair features help the network to achieve the top results.}
\label{fig:evaluations}
\end{figure*}
\vspace{-3mm}
\paragraph{Real Datasets}
We concentrate on real sets rather than synthetic ones and therefore our evaluations are against the diverse 3DMatch RGBD benchmark \cite{zeng20163dmatch}, in which 62 different real-world scenes retrieved from the pool of datasets Analysis-by-Synthesis \cite{valentin2016learning}, 7-Scenes \cite{shotton2013scene}, SUN3D \cite{xiao2013sun3d}, RGB-D Scenes v.2 \cite{lai2014unsupervised} and Halber et \cite{halber2016structured}. This collection is split into 2 subsets, 54 for training and validation, 8 for testing. The dataset typically includes indoor scenes like living rooms, offices, bedrooms, tabletops, and restrooms. See ~\cite{zeng20163dmatch} for details. As our input consists of only point geometry, we solely use the fragment reconstructions captured by Kinect sensor and not the color. 

\paragraph{Can PPFNet outperform the baselines on real data?}
We evaluate our method against hand-crafted baselines of Spin Images~\cite{johnson1999using}, SHOT~\cite{salti2014shot}, FPFH~\cite{rusu2009fast}, USC~\cite{tombari2010unique}, as well as 3DMatch \cite{zeng20163dmatch}, the state of the art deep learning based 3D local feature descriptor, the vanilla PointNet~\cite{qi2016pointnet} and CGF~\cite{Khoury_2017_ICCV}, a hybrid hand-crafted and deep descriptor designed for compactness. To set the experiments more fair, we also show a version of 3DMatch, where we use 2048 local patches per fragment instead of 5K, the same as in our method, denoted as 3DMatch-2K. We use the provided pre-trained weights of CGF~\cite{Khoury_2017_ICCV}. We keep the local patch size same for all methods. Our evaluation data consists of fragments from 7-scenes~\cite{shotton2013scene} and SUN3D~\cite{xiao2013sun3d} datasets. We begin by showing comparisons without applying RANSAC to prune the false matches. We believe that this can show the true quality of the correspondence estimator. Inspired by ~\cite{Khoury_2017_ICCV}, we accredit recall as a more effective measure for this experiment, as the precision can always be improved by better corresponding pruning~\cite{chin2015efficient,campbell2017globally}. Our evaluation metric directly computes the recall by averaging the number of matched fragments across the datasets:
\vspace{-0.5mm}
\begin{equation}
R = \frac{1}{M} \sum\limits_{s=1}^M \mathbbm{1} \Bigg( \Big[\frac{1}{|\Omega|} \sum\limits_{(i,j \in \Omega)} \mathbbm{1}\big( (\mathbf{x}_i - \mathbf{T}\mathbf{y}_j) < \tau_1\big)\Big] > \tau_2 \Bigg)
\end{equation}
where $M$ is the number of ground truth matching fragment pairs, having at least 30\% overlap with each other under ground-truth transformation $\mathbf{T}$ and $\tau_1=10cm$. $(i,j)$ denotes an element of the found correspondence set $\Omega$. $\mathbf{x}$ and $\mathbf{y}$ respectively come from the first and second fragment under matching. The inlier ratio is set as $\tau_2=0.05$.  
As seen from Tab. \ref{tab:3dmatchbenchmark}, PPFNet outperforms all the hand crafted counterparts in mean recall. It also shows consistent advantage over 3DMatch-2K, using an equal amount of patches. Finally and remarkably, we are able to show $\sim 2.7\%$ improvement on mean recall over the original 3DMatch, using only $\sim 40\%$ of the keypoints for matching. The performance boost from 3DMatch-2K to 3DMatch also indicates that having more keypoints is advantageous for matching. Our method expectedly outperforms both vanilla PointNet and CGF by $15\%$. 
We show in Tab.~\ref{tab:recallVSsample} that adding more samples brings benefit, but only up to a certain level ($<5$K). For PPFNet, adding more samples also increases the global context and thus, following the advent in hardware, we have the potential to further widen the performance gap over 3DMatch, by simply using more local patches. 
To show that we do not cherry-pick $\tau_2$ but get consistent gains,
we also plot the recall computed with the same metric for different inlier ratios in Fig. \ref{subfig:accuracy}. There, for the practical choices of $\tau_2$, PPFNet persistently remains above all others.
\begin{table}[h!]
    \small
    \centering
    \caption{Recall of 3DMatch for different sample sizes.}
    \setlength\tabcolsep{3pt}
    \begin{tabular}{lccccccccc}
    \toprule
    Samples & 128   & 256   & 512   & 1K    & 2K    & 5K    & 10K   & 20K   & 40K \\
    Recall & 0.24 & 0.32 & 0.40 & 0.47 & 0.51 & 0.59 & 0.59 & 0.56 & 0.60 \\
    \bottomrule
    \end{tabular}%
    \label{tab:recallVSsample}
\end{table}
\begin{table*}[htbp]
  \centering
  \caption{Our evaluations on the 3D-match benchmark after RANSAC. \textit{Kitchen} is from 7-scenes~\cite{shotton2013scene} and the rest from SUN3D~\cite{xiao2013sun3d}.}
  \resizebox{\textwidth}{!}{
    \begin{tabular}{lcccccccccccccccccc}
    \toprule
    \multicolumn{1}{c}{} & \multicolumn{2}{c}{Spin Images~\cite{johnson1999using}} & \multicolumn{2}{c}{SHOT~\cite{salti2014shot}} & \multicolumn{2}{c}{FPFH~\cite{rusu2009fast}} & \multicolumn{2}{c}{USC~\cite{tombari2010unique}} & \multicolumn{2}{c}{PointNet~\cite{qi2016pointnet}} & \multicolumn{2}{c}{CGF~\cite{Khoury_2017_ICCV}} & \multicolumn{2}{c}{3DMatch~\cite{zeng20163dmatch}} & \multicolumn{2}{c}{3DMatch-2K} & \multicolumn{2}{c}{PPFNet} \\
    \midrule
          & recall & prec. & recall & prec. & recall & prec. & recall & prec. & recall & prec. & recall & prec. & recall & prec. & recall & prec. & recall & prec. \\
    \midrule
    \multicolumn{1}{l}{Red Kitchen} & 0.27  & 0.49  & 0.21  & 0.44  & 0.36  & 0.52  & 0.52  & 0.60 & 0.76  & 0.60  & 0.72  & 0.54  & 0.85  & 0.72  & 0.80  & 0.54  & \textbf{0.90} & 0.66 \\
    \multicolumn{1}{l}{Home 1}      & 0.56  & 0.14  & 0.37  & 0.13  & 0.56  & 0.16  & 0.35  & 0.16 & 0.53  & 0.16  & 0.69  & 0.18  & \textbf{0.78} & 0.35  & 0.79  & 0.21  & 0.58  & 0.15 \\
    \multicolumn{1}{l}{Home 2}      & 0.35  & 0.10  & 0.30  & 0.11  & 0.43  & 0.13  & 0.47  & 0.24 & 0.42  & 0.13  & 0.46  & 0.12  & \textbf{0.61} & 0.29  & 0.52  & 0.14  & 0.57  & 0.16 \\
    \multicolumn{1}{l}{Hotel 1}     & 0.37  & 0.29  & 0.28  & 0.29  & 0.29  & 0.36  & 0.53  & 0.46 & 0.45  & 0.38  & 0.55  & 0.38  & \textbf{0.79} & 0.72  & 0.74  & 0.45  & 0.75  & 0.42 \\
    \multicolumn{1}{l}{Hotel 2}     & 0.33  & 0.12  & 0.24  & 0.11  & 0.36  & 0.14  & 0.20  & 0.17 & 0.31  & 0.18  & 0.49  & 0.15  & 0.59  & 0.41  & 0.60  & 0.22  & \textbf{0.68} & 0.22 \\
    \multicolumn{1}{l}{Hotel 3}     & 0.32  & 0.16  & 0.42  & 0.12  & 0.61  & 0.21  & 0.38  & 0.14 & 0.43  & 0.11  & 0.65  & 0.16  & 0.58  & 0.25  & 0.58  & 0.14  & \textbf{0.88} & 0.20 \\
    \multicolumn{1}{l}{Study Room}  & 0.21  & 0.07  & 0.14  & 0.07  & 0.31  & 0.11  & 0.46  & 0.17 & 0.48  & 0.16  & 0.48  & 0.16  & 0.63  & 0.27  & 0.57  & 0.17  & \textbf{0.68} & 0.16 \\
    \multicolumn{1}{l}{MIT Lab}     & 0.29  & 0.06  & 0.22  & 0.09  & 0.31  & 0.09  & 0.49  & 0.19 & 0.43  & 0.14  & 0.42  & 0.10  & 0.51  & 0.20  & 0.42  & 0.09  & \textbf{0.62} & 0.13 \\
    \midrule
    Average                         & 0.34  & 0.18  & 0.27  & 0.17  & 0.40  & 0.21  & 0.43  & 0.27 & 0.48  & 0.23  & 0.56  & 0.23 & 0.67  & 0.40  & 0.63  & 0.24  & \textbf{0.71} & 0.26 \\
    \bottomrule
    \end{tabular}%
  \label{tab:3dmatchbenchmark-ransac}%
  }
\end{table*}%
\vspace{-7mm}
\paragraph{Application to geometric registration}
Similar to~\cite{zeng20163dmatch}, we now use PPFNet in a broader context of transformation estimation. To do so, we plug all descriptors into the well established RANSAC based matching pipeline, in which the transformation between fragments is estimated by running a maximum of 50,000 RANSAC iterations on the initial correspondence set. We then transform the source cloud to the target by estimated 3D pose and compute the point-to-point error. This is a well established error metric~\cite{zeng20163dmatch}. Tab.~\ref{tab:3dmatchbenchmark-ransac} tabulates the results on the real datasets. Overall, PPFNet is again the top performer, while showing higher recall on a majority of the scenes and on the average. It is noteworthy that we always use 2048 patches, while allowing 3DMatch to use its original setting, 5K. Even so, we could get better recall on more than half of the scenes. When we feed 3DMatch 2048 patches, to be on par with our sampling level, PPFNet dominates performance-wise on most scenes with higher average accuracy. 
\vspace{-3mm}
\paragraph{Robustness to point density}
Changes in point density, a.k.a. sparsity, is an important concern for point clouds, as it can change with sensor resolution or distance for 3D scanners. This motivates us to evaluate our algorithm against others in varying sparsity levels. We gradually decrease point density on the evaluation data and record the accuracy. Fig.~\ref{subfig:sparsity} shows the significant advantage of PPFNet, especially under severe loss of density (only $6.5\%$ of points kept). Such robustness is achieved due to the PointNet backend and the robust point pair features. 
\begin{table}[t!]
  \centering
  \caption{Average per-patch runtime of different methods.}
  \resizebox{\columnwidth}{!}{
  \small
    \begin{tabular}{lccc}
    \toprule
          & input preparation & inference / patch & total \\
    \midrule
    3DMatch & $0.31ms$ on GPU & $2.9ms$ on GPU & $3.21ms$ \\
    PPFNet  & ${2.24ms}$ {on CPU} & ${55\mu s}$ {on GPU} & $\mathbf{2.25ms}$ \\
    \bottomrule
    \end{tabular}
  \label{tab:runtime}%
  }
  \vspace{-3mm}
\end{table}%
\vspace{-3mm}
\paragraph{How fast is PPFNet?}
We found PPFNet to be lightning fast in inference and very quick in data preparation since we consume a very raw representation of data. Majority of our runtime is spent in the normal computation and this is done only once for the whole fragment. The PPF extraction is carried out within the neighborhoods of only 2048 sample points. Tab. \ref{tab:runtime} shows the average running times of different methods and ours on an NVIDIA TitanX Pascal GPU supported by an Intel Core i7 3.2GhZ 8 core CPU. Such dramatic speed-up in inference is enabled by the parallel-PointNet backend and our simultaneous correspondence estimation during inference for all patches. Currently, to prepare the input for the network, we only use CPU, leaving GPU idle for more work. This part can be easily implemented on GPU to gain even further speed boosts.
\subsection{Ablation Study}
\paragraph{N-tuple loss} 
We train and test our network with 3 different losses: contrastive (pair)~\cite{hadsell2006dimensionality}, triplet~\cite{hoffer2015deep} and our N-tuple loss on the same dataset with identical network configuration. Inter-distance distribution of correspondent pairs and non-correspondent pairs are recorded for the train/validation data respectively. Empirical results in Fig. \ref{fig:loss_compare} show that the theoretical advantage of our loss immediately transfers to practice: Features learned by N-tuple are better separable, i.e. non-pairs are more distant in the embedding space and pairs enjoy a lower standard deviation. N-tuples loss repels non-pairs further in comparison to contrastive and triplet losses because of its better knowledge of global correspondence relationships.  Our N-tuple loss is general and thus we strongly encourage the application also to other domains such as pose estimation~\cite{wohlhart2015learning}.
\insertimageC{1}{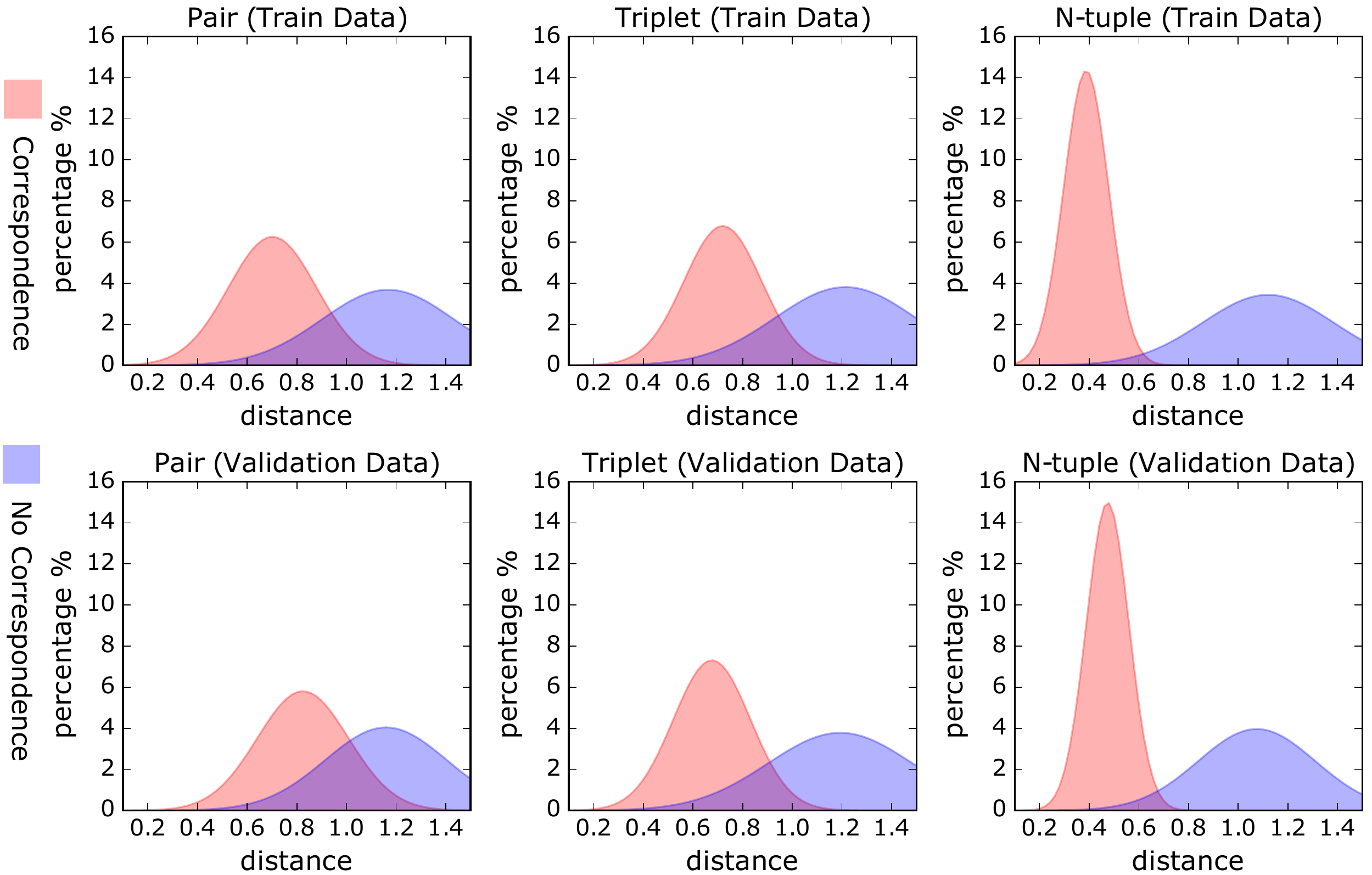}{N-tuple Loss (c) lets the PPFNet better separate the matching vs non-matching pairs w.r.t. the traditional contrastive (a) and triplet (b) losses.}{fig:loss_compare}{t!}
\begin{table}[t!]
  \centering
  \caption{Effect of different components in performance: Values depict the number of correct matches found to be $5\%$ inlier ratio.}
  \setlength\tabcolsep{13.5 pt}
  \small
    \begin{tabular}{lcc}
    \toprule
    Method & Train & Validation \\
    \midrule
    Without points and normals & 0\%     & 6\% \\
    Vanilla PointNet~\cite{qi2016pointnet} & 47\%    & 41\% \\
    Without global context & 48\%    & 46\% \\
    Without PPF & 65\%    & 48\% \\
    \textbf{All combined}   & \textbf{67\%}    & \textbf{56\%} \\
    \bottomrule
    \end{tabular}%
    \vspace{-3mm}
  \label{tab:ppf_boost}%
\end{table}%
\insertimageStar{1}{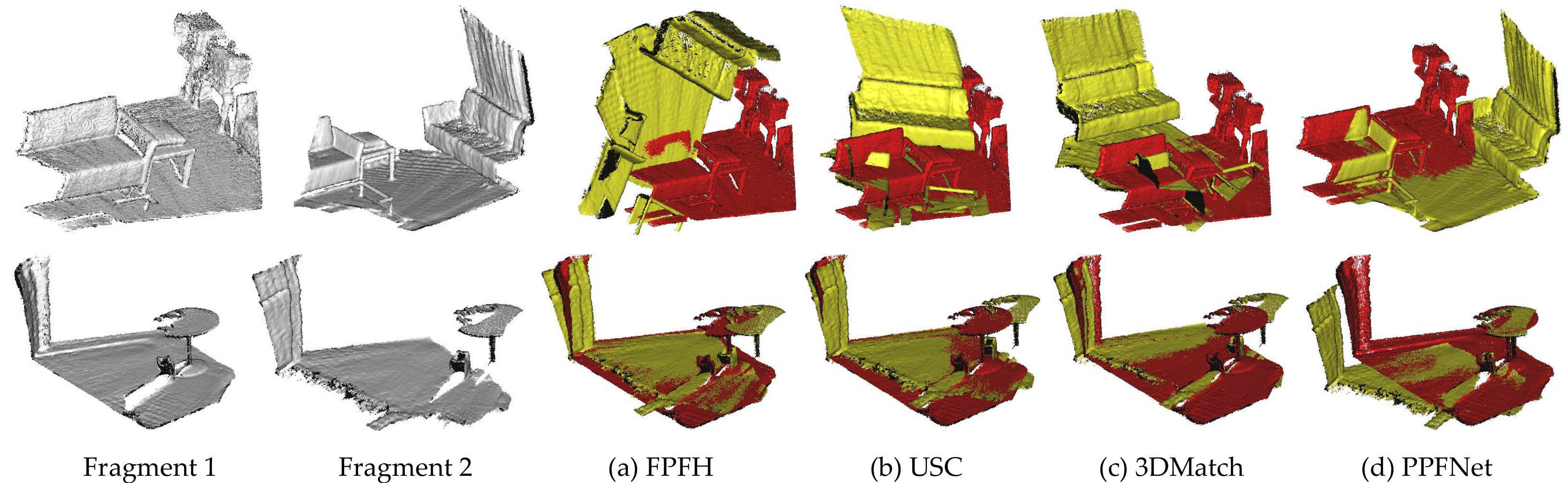}{Visualization of estimated transformations. Thanks to its robustness and understanding of global information, PPFNet can operate under challenging scenarios with confusing, repetitive structures as well as mostly planar scenes with less variation in geometry.}{fig:matchingvisuals}{t!}
\vspace{-3mm}
\paragraph{How useful is global context for local feature extraction?} We argue that local features are dependent on the context. A corner belonging to a dining table should not share the similar local features of a picture frame hanging on the wall. A table is generally not supposed to be attached vertically on the wall. To assess the returns obtained from adding global context, we simply remove the global feature concatenation, keep the rest of the settings unaltered, and re-train and test on two subsets of pairs of fragments. Our results are shown in Tab. \ref{tab:ppf_boost}, where injecting global information into local features improves the matching by $18\%$ in training and $7\%$ in validation set as opposed to our baseline version of Vanilla PointNet~\footnote{Note that this doesn't $100\%$ correspond to the original version, as we modified PointNet with task specific losses for our case.}, which is free of global context and PPFs. Such significance indicates that global features aid discrimination and are valid cues also for local descriptors.
\insertimageC{1}{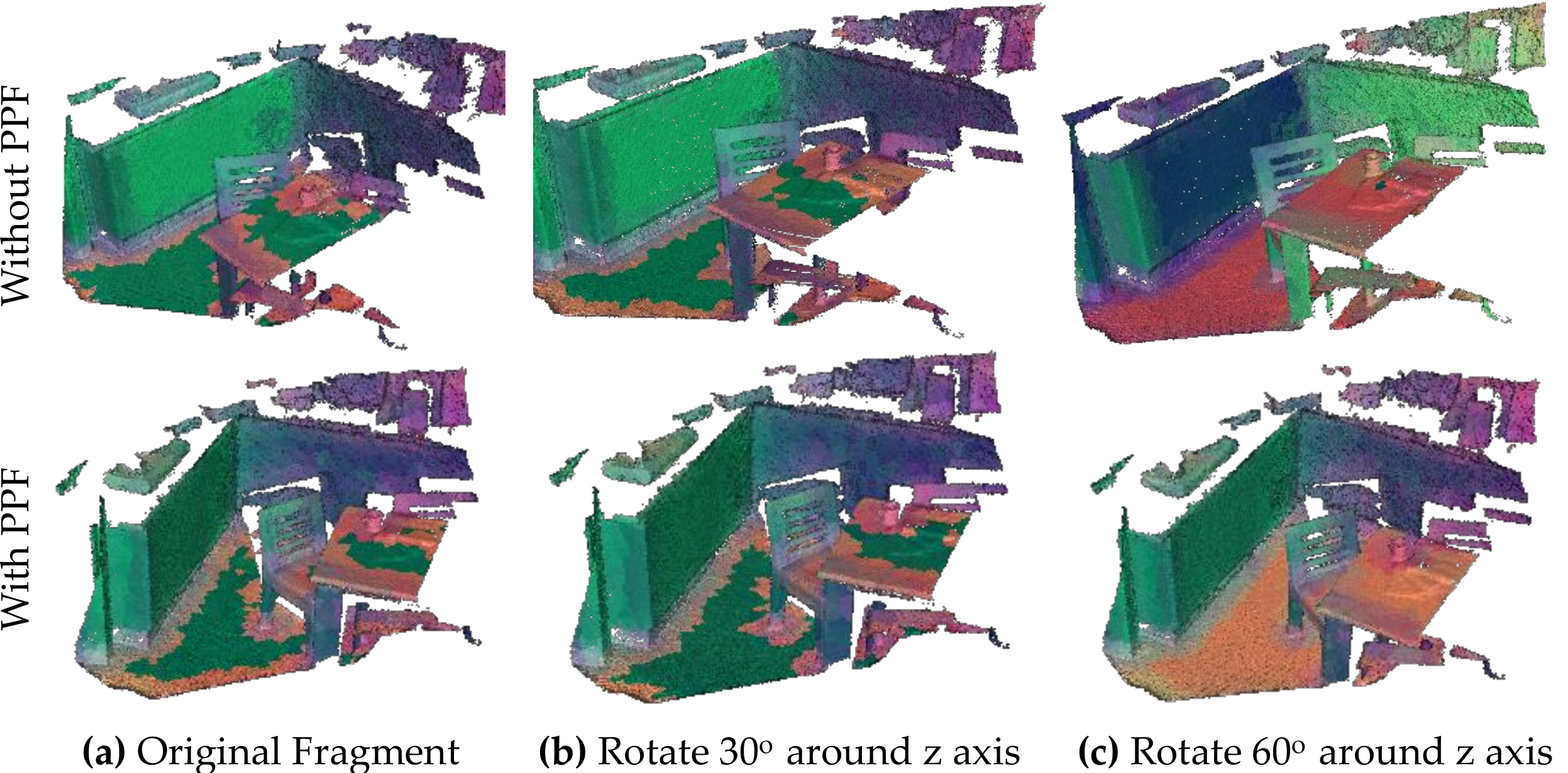}{Inclusion of PPF makes the network more robust to rotational changes as shown, where the appearance across each row is expected to stay identical, for a fully invariant feature.}{fig:ppf_rotation}{t!}
\vspace{-5mm}
\paragraph{What does adding PPF bring?}
We now run a similar experiment and train two versions of our network, with/without incorporating PPF into the input. The contribution is tabulated in Tab. \ref{tab:ppf_boost}. There, a gain of $1\%$ in training and $5\%$ in validation is achieved, justifying that inclusion of PPF increases the discriminative power of the final features. 

While being a significant jump, this is not the only benefit of adding PPF. Note that our input representation is composed of $33\%$ rotation-invariant and $66\%$ variant representations. This is already advantageous to the state of the art, where rotation handling is completely left to the network to learn from data. We hypothesize that an input guidance of PPF would aid the network to be more tolerant to rigid transformations. To test this, we gradually rotate fragments around z-axis to 180$^\circ$ with a step size of 30$^\circ$ and then match the fragment to the non-rotated one.  As we can observe from Tab. \ref{tab:ppf_rotation}, with PPFs, the feature is more robust to rotation and the ratio in matching performance of two networks opens as rotation increases.  In accordance, we also show a visualization of the descriptors at Fig. \ref{fig:ppf_rotation} under small and large rotations. To assign each descriptor an RGB color, we use PCA projection from high dimensional feature space to 3D color space by learning a linear map~\cite{Khoury_2017_ICCV}. It is qualitatively apparent that PPF can strengthen the robustness towards rotations. All in all, with PPFs we gain both accuracy and robustness to rigid transformation, the best of seemingly contradicting worlds. It is noteworthy that using only PPF introduces full invariance besides the invariance to permutations and renders the task very difficult to learn for our current network. We leave this as a future challenge. 

\begin{table}[t!]
  \centering
  \caption{Effect of point pair features in robustness to rotations.}
  \resizebox{\columnwidth}{!}{
    \begin{tabular}{lccccccc}
    \toprule
    z-rotation & $0^\circ$    & $30^\circ$   & $60^\circ$   & $90^\circ$   & $120^\circ$  & $150^\circ$  & $180^\circ$ \\
    \midrule
    with PPF & 100.0\% & 53.3\% & 35.0\% & 20.0\% & 8.3\% & 5.0\% & 0.0\% \\
    w/o PPF & 100.0\% & 38.3\% & 23.3\% & 11.7\% & 1.7\% & 0.0\% & 0.0\% \\
    \bottomrule
    \end{tabular}%
    }
    \vspace{-3mm}
  \label{tab:ppf_rotation}%
\end{table}%
A major limitation of PPFNet is quadratic memory footprint, limiting the number of used patches to 2K on our hardware. This is, for instance, why we cannot outperform 3DMatch on fragments of \textit{Home-2}. With upcoming GPUs, we expect to reach beyond 5K, the point of saturation.
\vspace{-3mm}
\section{Conclusion}
\vspace{-1mm}
We have presented \textbf{PPFNet}, a new 3D descriptor tailored for point cloud input. By generalizing the contrastive loss to N-tuple loss to fully utilize available correspondence relatioships and retargeting the training pipeline, we have shown how to learn a globally aware 3D descriptor, which outperforms the state of the art not only in terms of recall but also speed. Features learned from our PPFNet is more capable of dealing with some challenging scenarios, as shown in Fig. \ref{fig:matchingvisuals}. Furthermore, we have shown that designing our network suitable for set-input such as point pair features are advantageous in developing invariance properties. 

Future work will target memory bottleneck and solving the more general rigid graph matching problem.

\clearpage

{\small
\bibliographystyle{ieee}
\bibliography{egbib}
}


\setcounter{section}{0}
\renewcommand\thesection{\Alph{section}}
\newcommand{\suppsection}{\subsection}
\clearpage
\section{Appendix}
\subsection{Further Architectural Details}
\begin{figure*}[b!]
\twocolumn[
{\includegraphics[width=\linewidth]{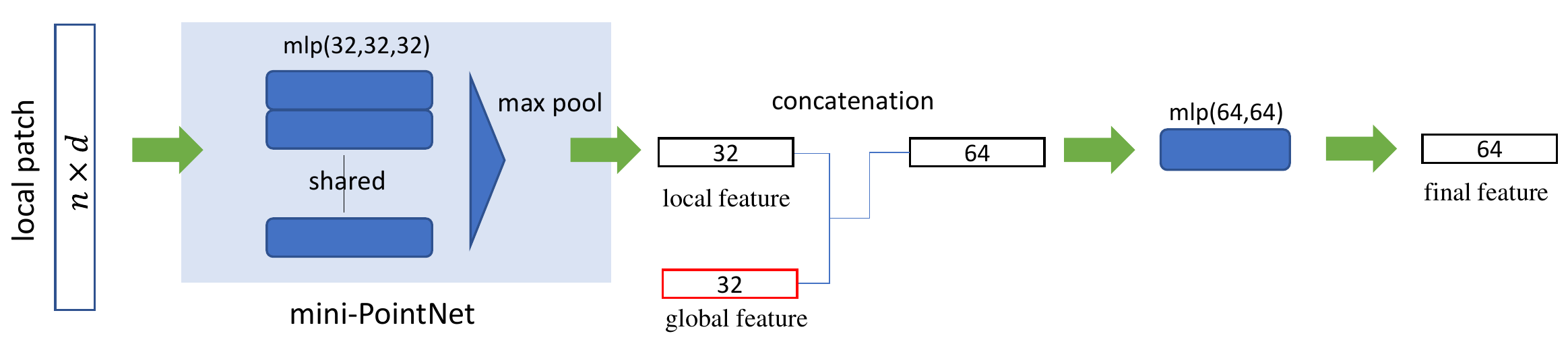}
\caption{Pipeline for processing a single local patch.}\par\bigskip}]
\label{fig:small_net}
\end{figure*}

Due to the constraint of GPU memory, we adopt a minimized version of vanilla PointNet in our implementation. Fig. \ref{fig:small_net} demonstrates a pipeline for processing one single patch and more details of our network. 

The size of a local patch is $n \times d$, where $n=1024$ is number of points in the patch and $d$ depends on the specific representation of local patch. For only point coordinates and normals, $d=6$; for the one with extra PPF, $d=10$.

A patch is first sent into a mini-PointNet with three layers, each has 32 nodes, and then a max pooling function aggregates all the information into a 32-dimensional local feature. After combining with the 32-dimensional global feature, it is further processed by a two-layer MLP, in which each layer has 64 nodes. The dimension of the final feature for the local patch is 64.

Fig. \ref{fig:res_registration} demonstrates some qualitative fragment registration results in Section 5 in the paper, showing that learned features by PPFNet are able to cope with challenging point cloud matching  problems under different situations.

\subsection{Algorithmic Details}
\paragraph{Sampling algorithm} 
How we sample the point cloud down to 2048 keypoints (samples) plays an important role in learning. We try to be spatially as uniformly distributed as possible so that the samples are further apart and less dependent on one another. For this, we use the greedy algorithm given in Algorithm 1 inspired by ~\cite{birdal2017sampling}. Note that the algorithm involves a search over the so-far-sampled cloud, which we speed up using a voxel-grid.

\paragraph{Normal computation} 
To achieve speed-up in point pair feature calculation, we pre-compute the normals of the input fragment. To compute a normal, the tangent plane to each local neighborhood is approximated by the least-square fitting as proposed in~\cite{Hoppe1992}. Computing the equation of the plane then boils down to a analysis of a covariance matrix created from the nearest neighbors:
\begin{equation}
\mathbf{C} = \frac{1}{n} \sum\limits_{i=1}^n (\mathbf{x}_i-\bar{\mathbf{x}})(\mathbf{x}_i-\bar{\mathbf{x}})^T
\end{equation}
where $\bar{\mathbf{x}}$ denotes the mean, or the center of the local patch. The equation of the plane is then computed from the eigenvectors of $\mathbf{C}$. Due to the sign ambiguity in eigenvector analysis, the direction of the resulting normal is unknown. Thus, we use the convention where each surface normal is flipped towards the camera by ensuring the dot product between the viewpoint vector and surface normal be acute: $ -\mathbf{p}\cdot \mathbf{n} < {\pi}/{2}$.
\setcounter{algorithm}{-1}
\begin{algorithm}[htbp]
	\footnotesize
	\caption{Uniform Sampling}
	\label{alg:sampling}
	\begin{algorithmic}
		\Require{Source point cloud $\mathbf{X}$, Relative threshold $\tau$}
		\Ensure{Sampled point cloud $\mathbf{S}$ and its normals $\mathbf{N}$}
		\State $\text{Compute normals for $\mathbf{X}$}$
		\State $\mathbf{S} \gets []$
		\State $\mathbf{N} \gets []$
		\State $R_d \gets diameter(\mathbf{X})$
		\For {$\mathbf{x} \in \mathbf{X}$}
		\State $d_{min} = \min_{(\mathbf{t} \in \mathbf{S})} |\mathbf{x}-\mathbf{t}|$
		\If {$(d_{min} > \tau R_d)$}
		\State $\mathbf{S} \gets \mathbf{S} \bigcup \mathbf{x}$
		\State $\mathbf{N} \gets \mathbf{N} \bigcup \mathbf{n}(\mathbf{x})$ \Comment sample the normal as well
		\EndIf
		\EndFor
		\caption{Distance Constrained Sampling}
	\end{algorithmic}
\end{algorithm}
\insertimageStar{0.9}{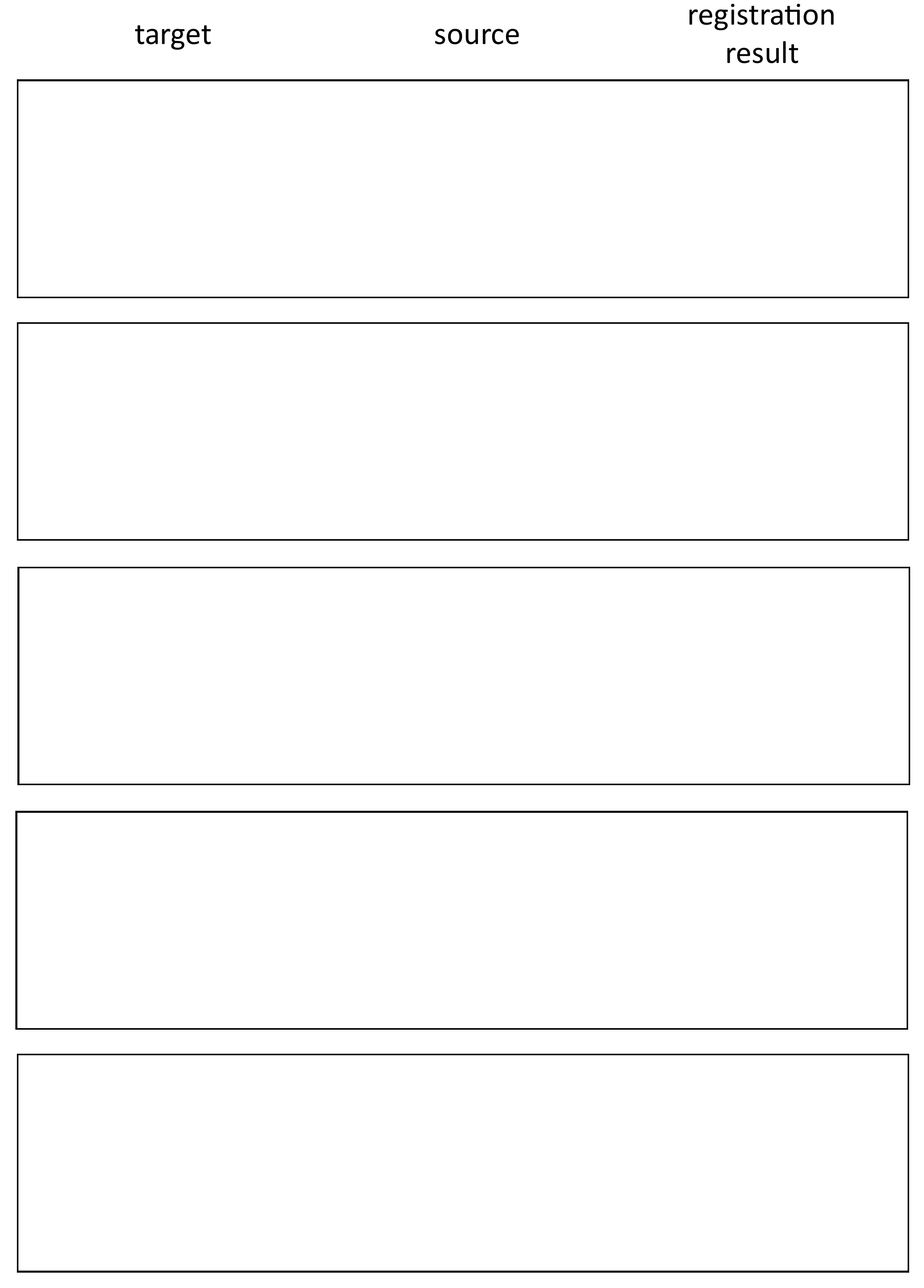}{Qualitative registration results of 5 fragment pairs}{fig:res_registration}{b!}

\end{document}